# A Closed Form Solution to Multi-View Low-Rank Regression

## Shuai Zheng, Xiao Cai, Chris Ding, Feiping Nie, Heng Huang

Department of Computer Science and Engineering
University of Texas at Arlington, TX, USA
zhengs123@gmail.com, xiao.cai@mavs.uta.edu, chqding@uta.edu,
feipingnie@gmail.com, heng@uta.edu

## Abstract

Real life data often includes information from different channels. For example, in computer vision, we can describe an image using different image features, such as pixel intensity, color, HOG, GIST feature, SIFT features, etc.. These different aspects of the same objects are often called multi-view (or multi-modal) data. Low-rank regression model has been proved to be an effective learning mechanism by exploring the low-rank structure of real life data. But previous low-rank regression model only works on single view data. In this paper, we propose a multi-view low-rank regression model by imposing low-rank constraints on multi-view regression model. Most importantly, we provide a closed-form solution to the multi-view low-rank regression model. Extensive experiments on 4 multi-view datasets show that the multi-view low-rank regression model outperforms single-view regression model and reveals that multi-view low-rank structure is very helpful.

## Introduction

In many tasks, a single object can be described using information from different channels (or views). For example, a 3-D object can be described using pictures from different angles; a website can be described using the words it contains, and the hyperlinks it contains; an image can be described using different features, such as SIFT feature, and HOG feature; in daily life, a person can be characterized using age, height, weight and so on. These data all comes from different aspects and channels. Multi-view problems aim to improve existing single view model by learning a model utilizing data collected from multiple channels (Rüping and Scheffer 2005) (de Sa 2005) (Zhou and Burges 2007).

Low-rank regression model has been proved to be an effective learning mechanism by exploring the low-rank structure of real life data (Xiang et al. 2012) (Evgeniou and Pontil 2007) (Cai et al. 2013). Existing regression models only work on single view data. To be specific, linear regression finds a linear model with respect to the single view feature data to fit target class data (Seber and Lee 2012). Let matrix $B \in \Re^{p \times c}$ be the parameter of the linear model. Linear regression solves a problem of $\min_B ||Y - X^T B||_F^2$, where $X = [\mathbf{x}_1, \mathbf{x}_2, ..., \mathbf{x}_n] \in \Re^{p \times n}$ is the single view feature data matrix and $Y \in \Re^{n \times c}$ is the target class indicator matrix. Ridge regression can achieve better results by adding a Frobenius norm based regularization on linear regression loss objective (Hoerl and Kennard 1970) (Marquaridt 1970). Ridge regression solves the problem $\min_B ||Y - X^T B||_F^2 + \lambda ||B||_F^2$, where $\lambda$ is the regularization weight parameter. Cai (Cai et al. 2013) showed that when $B$ is low-rank, regression is equivalent to linear discriminant analysis based regressions. However, all these work only works for single-view problems.

In this paper, we propose a multi-view low-rank regression model by imposing low-rank constraints on regression model. This model can be solved using closed-form solution directly. In linear regression, low rank parameter matrix $B^\nu$ is dependent on view $\nu$. Through theoretical analysis, we show that multi-view low-rank regression model is equivalent to do regression in the subspace of each view. In other words, let $B^\nu = A_\nu B$, and it is equivalent to find the shared regression parameter matrix $B$ under the subspace transformation $A_\nu$ with respect to view $\nu$. Extensive experiments performed on 4 multi-view datasets show that the proposed model outperforms single-view regression model and reveals that low-rank structure can improve the classification result of a full-rank model.

**Notations.** In this paper, matrices are written in uppercase letters, such as $X$, $Y$. Vectors are written in bold lower case letters, such as $\mathbf{x}$, $\mathbf{y}$. $\text{Tr}(X)$ means the trace operation for matrix $X$.

## Multi-view Low Rank Regression

Assume that there are $v$ views and $c$ classes, $p_\nu$ is the dimension of view $\nu$, $n_j$ is the sample size of the $j$-th class, and $n$ is the total sample size. Let $X_\nu = [\mathbf{x}_1^\nu, ..., \mathbf{x}_n^\nu] \in \Re^{p_\nu \times n}$ be the data matrix of view $\nu$, $\nu = 1, 2, ..., v$, and $Y = [\mathbf{y}_1, ..., \mathbf{y}_c] \in \Re^{n \times c}$ is the normalized class indicator matrix, i.e. $Y_{ij} = 1/\sqrt{n_j}$ if the $i$-th data point belongs to the $j$-th class and $Y_{ij} = 0$ otherwise.

We try to minimize the residual of low rank regression model in each class and in each view. Loss function of multi-view low rank ridge regression can be proposed as in Eq.(1):

$$J_0 = \sum_{\nu=1}^{v} \sum_{k=1}^{c} \{\|\mathbf{y}_k - (X_\nu^T \beta_k^\nu + f_k^\nu \mathbf{e})\|_2^2 + \lambda_\nu \|\beta_k^\nu\|_2^2\}$$
$$= \sum_{\nu=1}^{v} \{\|Y - (X_\nu^T B^\nu + EF^\nu)\|_F^2 + \lambda_\nu \|B^\nu\|_F^2\} \quad (1)$$

where projection matrix $B^\nu = [\beta_1^\nu, ..., \beta_c^\nu] \in \Re^{p_\nu \times c}$, bias $F^\nu = diag(f_1^\nu, ..., f_c^\nu)$, $E = [\mathbf{e}, ..., \mathbf{e}] \in \Re^{n \times c}$. $\mathbf{e}$ is a $n$-dimensional column vector with all elements equal to 1. $\lambda_\nu$ is the regularization parameter of view $\nu$. Let's introduce low rank projection $B^\nu$ with rank $s$, $s < min(p_\nu, c)$,

$$\beta_k^\nu = A_\nu b_k, \quad \text{or} \quad B^\nu = A_\nu B, \quad (2)$$

where $A_\nu \in \Re^{p_\nu \times s}$, and $B = (\mathbf{b}_1, ..., \mathbf{b}_c) \in \Re^{s \times c}$. Therefore, the objective function Eq.(1) can be written as:

$$J_1 = \sum_{\nu=1}^{v} \{\|Y - (X_\nu^T A_\nu B + EF^\nu)\|_F^2 + \lambda_\nu \|A_\nu B\|_F^2\} \quad (3)$$

It is noteworthy that from Eq.(3), we can see that multi-view low-rank regression model is equivalent to do regression in the subspace of each view. Matrix $A_\nu$ is the subspace matrix of view $\nu$. Matrix $B$ is the shared regression parameter matrix of all views.

### Closed form solution

We now present the closed form solution of the Multi-view Low Rank Regression. Before we talk about the closed form solution, we present Lemma 1 to simplify Eq.(3).

**Lemma 1.** *The bias $f_k^\nu$ can be solved and eliminated from $J_1$, which is thus simplified into*

$$J_1 = \sum_{\nu=1}^{v} \{\|Y^c - X_\nu^{cT} A_\nu B\|_F^2 + \lambda_\nu \|A_\nu B\|_F^2\} \quad (4)$$

*where bias $f_k^\nu$ relates to $B$ as*

$$f_k^{\nu*} = \bar{y}_k - \bar{\mathbf{x}}_\nu^T A_\nu b_k^\nu \quad (5)$$

*and $X_\nu^c = X_\nu - \bar{\mathbf{x}} \mathbf{e}^T$ is centered data matrix of view $\nu$ and $Y^c = Y - (\bar{y}_1, ..., \bar{y}_c)\mathbf{e}$ is centered class indicator matrix.*

*Proof.* Taking derivative of Eq.(3) w.r.t. $f_k^\nu$ and setting it to zero, the optimal solution of $f_k^\nu$ is given as in Eq.(5), where $\bar{y}_k$ is a real number, $\bar{y}_k = \sum_{i=1}^{n} y_{ki}/n$, $\bar{\mathbf{x}}_\nu = \sum_{i=1}^{n} \mathbf{x}_i^\nu/n \in \Re^{p_\nu \times 1}$. Substituting Eq.(5) into Eq.(3), we have Eq.(4). □

In the rest of this paper, we focus on solving Eq.(4). For simplicity of notations, we drop $c$ in $X_\nu^c$ and use $X_\nu$ to denote the centered $X_\nu$. Similarly, we drop $c$ in $Y^c$ and use $Y$ to denote the centered $Y$.

Now we present Theorem 1 to give the closed form solution of multi-view low-rank regression model.

**Theorem 1.** *The optimal solution of $J_1(\{A_\nu\}, B)$ is the following:*

*1. $\{A_\nu\}$ is given by the optimal solution of the following problem:*

$$\max_{\{A_\nu\}} Tr(G^{-1} H Y Y^T H^T) \quad (6)$$

*where*

$$G = G(\{A_\nu\}) \triangleq \sum_{\nu} A_\nu^T (X_\nu X_\nu^T + \lambda_\nu I) A_\nu, \quad (7)$$

$$H = H(\{A_\nu\}) \triangleq \sum_{\nu} A_\nu^T X_\nu \quad (8)$$

*2. $B$ is given by*

$$B^* = G^{-1} H. \quad (9)$$

*Proof.* Taking derivative of Eq.(4) w.r.t. $B$, we have

$$\frac{\partial J}{\partial B} = -2 \sum_{\nu} A_\nu^T X_\nu Y + 2 \sum_{\nu} A_\nu^T X_\nu X_\nu^T A_\nu B$$
$$+ 2\lambda_\nu \sum_{\nu} A_\nu^T A_\nu B. \quad (10)$$

Setting Eq.(10) to zero, we have Eq.(9).
Substituting Eq.(9) in Eq.(4), we have

$$J = -\min_{\{A_\nu\}} Tr(G^{-1} H Y Y^T H^T) \quad (11)$$

where $G = G(\{A_\nu\}) \triangleq \sum_\nu A_\nu^T (X_\nu X_\nu^T + \lambda_\nu I) A_\nu$, $H = H(\{A_\nu\}) \triangleq \sum_\nu A_\nu^T X_\nu$. Eq.(11) is equivalent to Eq.(6). □

Furthermore, we present Theorem 2 to give the closed form solution for Eq.(6). Let

$$A = \begin{pmatrix} A_1 \\ A_2 \\ ... \\ A_v \end{pmatrix}, \qquad X = \begin{pmatrix} X_1 \\ X_2 \\ ... \\ X_v \end{pmatrix}, \quad (12)$$

$$S_b = XYY^T X^T, \quad (13)$$

$$S_t = diag(X_1 X_1^T + \lambda_1 I, ..., X_v X_v^T + \lambda_v I), \quad (14)$$

**Theorem 2.** *Eq.(6) is equivalent to*

$$\max_A Tr[(A^T S_t A)^{-1} A^T S_b A], \quad (15)$$

*where the optimal solution $A^*$ is given by eigenvectors of $S_t^{-1} S_b$ that correspond to the $s$ largest eigenvalues.*

### Algorithm

We present Algorithm 1 to summarize the steps of multi-view low-rank regression model. One of the advantages of our model is that it can be solved using closed-form solution directly. The input of this algorithm is (1) centered and normalized data matrix $X_\nu \in \Re^{p_\nu \times n}$ from view $\nu$, where $\nu = 1, 2, ..., v$, $v$ is view number, $p_\nu$ is the dimension of view $\nu$ and $n$ is sample number, (2) class indicator matrix $Y \in \Re^{n \times c}$, (3) regularization weight parameter $\lambda_\nu$, (4) rank $s$, which is less than the class number $c$. The output of this algorithm is matrix $A_\nu \in \Re^{p_\nu \times s}$ and $B \in \Re^{s \times c}$. We can compute $S_b$ and $S_t$ using Eq.(13) and Eq.(14). In step 2, we compute $A$, which is those eigenvectors of $S_t^{-1} S_b$ that correspond to the $s$ largest eigenvalues. We should use Eq.(12) to restore $A_\nu$ from $A$. Finally, we compute $B$ using Eq.(9).

**Algorithm 1** Multi-view low-rank regression

**Input:** Data matrix $X_\nu \in \Re^{p_\nu \times n}$, class indicator matrix $Y \in \Re^{n \times c}$, regularization weight parameter $\lambda_\nu$, rank $s < c$, $\nu = 1, 2, ..., v$
**Output:** Matrix $A_\nu \in \Re^{p_\nu \times s}$ and $B \in \Re^{s \times c}$, $\nu = 1, 2, ..., v$
1: Compute $S_b$ and $S_t$ using Eq.(13) and Eq.(14)
2: Compute $A_\nu$ using the optimal solution of Eq.(15)
3: Compute $B$ using Eq.(9)

Table 1: Multi-view datasets attributes.

| Data | $n$ | $c$ | $v$ | $p_\nu$ |
|---|---|---|---|---|
| MSRC | 210 | 7 | 4 | 1302, 512, 100, 256 |
| Caltech | 1230 | 20 | 4 | 1302, 512, 100, 256 |
| Cornell | 195 | 5 | 3 | 107, 20, 15 |
| Cora | 2708 | 7 | 3 | 101, 180, 75 |

## Multi-view Full Rank Regression

Low-rank regression model has been proved to be an effective learning mechanism by exploring the low-rank structure of real life data. Will the multi-view low-rank regression model be able to capture the low-rank structure and improve the performance of a full-rank model? We will compare the performance of multi-view low-rank regression model with a full-rank model in experiment section.

In the case of multi-view full-rank regression, rank $s = c$, there is no constraint on $B^\nu$ in Eq.(1) and we will not use Eq.(2). To be specific, we will minimize the objective Eq.(4):

$$J_1 = \sum_{\nu=1}^{v} \{\|Y - X_\nu^T B^\nu\|_F^2 + \lambda_\nu \|B^\nu\|_F^2\} \quad (16)$$

Eq.(16) can be solved using close form solution. Taking derivative of Eq.(16) w.r.t. $B^\nu$ and setting it to zero, the optimal solution of $B^\nu$ is given as

$$B^\nu = (X_\nu X_\nu^T + \lambda_\nu I)^{-1} X_\nu Y, \quad (17)$$

where $I \in \Re^{p_\nu \times p_\nu}$ is an identity matrix.

## Connections to other Multi-view work

Various multi-view learning models have been studied and all multi-view models are expected to have better performance than single view models. Existing multi-view approaches mainly are inspired from spectral clustering and subspace learning. de Sa (de Sa 2005) developed a spectral clustering algorithm for only two views by creating a bipartite graph based on the "minimizing-disagreement" idea. Zhou (Zhou and Burges 2007) developed a multi-view spectral clustering model via generalizing the single view normalized cut to the multi-view case. They try to find a cut which is close to be optimal on each single-view graph by exploiting a mixture of Markov chains associated with graphs of different views. Kumar (Kumar and Daumé 2011) proposed a co-training flavour spectral clustering algorithm and use spectral embedding from one view to constrain the similarity graph used for the other view. Kumar (Kumar, Rai, and Daume 2011) used the philosophy of co-regularization, which has been used in the past for semi-supervised learning problems, to make the clusterings in different views agree with each other.

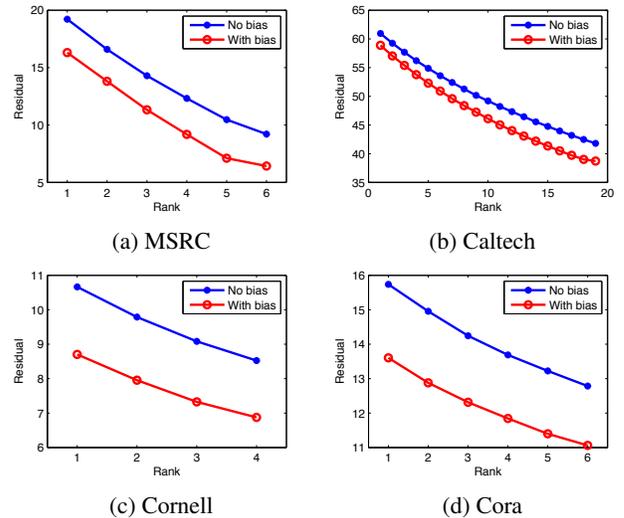

(a) MSRC  (b) Caltech

(c) Cornell  (d) Cora

Figure 1: Effect of regression bias in Eq.(1) or Eq.(3).

Multi-view learning models from the point of view of subspace learning mainly try to find a subspace for each view and then develop a learning model across views in their subspaces. Canonical-Correlation Analysis (CCA) (Hotelling 1936) was first used to study the correlation of two views in their respective subspaces. Hardoon (Hardoon, Szedmak, and Shawe-Taylor 2004) (Hardoon and Shawe-Taylor 2009) designed an Kernel Canonical-Correlation Analysis to extract patterns from two views. Chaudhuri (Chaudhuri et al. 2009) proposed a CCA-based subspace multi-view learning approach to find a subspace such that the objects of different classes are well-separated and within-class distance is minimized. Greene (Greene and Cunningham 2009) developed a Non-negative Matrix Factorization (NMF) (Lee and Seung 1999) approach to effectively identify common patterns and reconcile between-view disagreements by combining data from multiple views.

The proposed multi-view low-rank regression model should be categorized into the class of subspace learning multi-view. The important contribution of this paper is that we developed low-rank regression model to study multi-view problems. Surprisingly, there exists closed form solution to multi-view low-rank regression model.

## Experiments

In this section, we perform extensive experiments on 4 multiple-view datasets. Through model learning, we systematically explore the best settings of regression bias, regularization weight parameter $\lambda_\nu$ and how to do classification using multi-view regression. We compare the classification accuracy of multi-view low-rank ridge regression with single-view regression, linear regression and full rank ridge regression.

1975

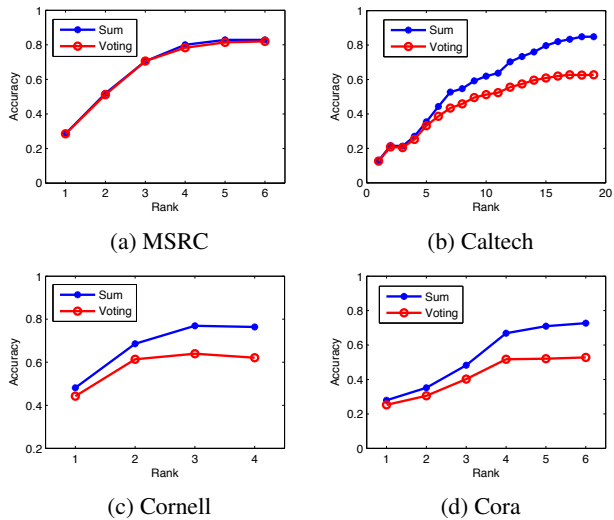

Figure 2: Classification using different voting or sum methods.

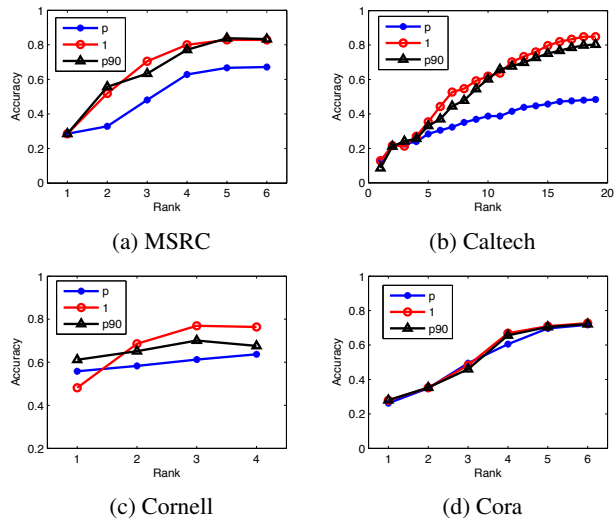

Figure 3: Regularization weight parameter $\lambda_\nu$.

## Datasets

Various multi-view datasets are used. These datasets include image datasets MSRC (Lee and Grauman 2009) and Caltech (Fei-Fei, Fergus, and Perona 2007), website dataset Cornell (Craven et al. 2000) and scientific publication dataset Cora (McCallum et al. 1999). Cornell and Cora are downloaded from (Grimal 2014). Summary of the datasets attributes are presented in Table 1, where $n$ is sample number, $c$ is class number, $v$ is view number and $p_\nu$ lists the dimensions of different views.

**MSRC** is an image scene data, including trees, buildings, planes, cows, faces, cars and so on. It has 210 images from 7 classes. We extract different features from this data. The 4 views we used in this paper are CENTRIST(1302 dimensions), GIST (512 dimensions), HOG (100 dimensions) and LBP (256 dimensions).

**Caltech** is a subset of Caltech 101 image data. It has images from 20 classes, including Faces, Leopards, Motorbikes, binocular, Brain, Camera, etc.. This data has 1230 images and 4 features are extracted from this data, including CENTRIST(1302 dimensions), GIST (512 dimensions), HOG (100 dimensions) and LBP (256 dimensions).

**Cornell** contains 195 documents over the 5 types (student, project, course, staff, faculty). There exists referral links among these documents. We use 3 views to describe the same document, including content view (107 dimensions), inbound-link view (20 dimensions) and outbound-link view (15 dimensions).

**Cora** consists of 2708 scientific publications classified into one of seven classes (Neural Networks, Rule Learning, Reinforcement Learning, Probabilistic Methods, Theory, Genetic Algorithms, Case Based). The citation network consists of links among those publications. The 3 views used in our experiments include content view (101 dimensions), inbound-link view (180 dimensions) and outbound-link view (75 dimensions).

## Model learning

Through model learning, we systematically explore the best settings of regression bias, regularization weight parameter $\lambda_\nu$ and how to do classification using multi-view regression.

**Effect of regression bias** To validate that adding bias to regression will reduce fitting residual, Figure 1 compares the residual of class indicator matrix $Y$ using two $f_k^\nu$ values: (1). using Eq.(5), denoted by "With bias" line (red circle line), (2) $f_k^\nu = 0$, denoted by "No bias" line (blue dot line). Residual $r$ is defined as

$$r = \sum_{\nu=1}^{v} \|Y - X_\nu^T A_\nu B\|_F^2. \qquad (18)$$

$r$ is the summation of label matrix residuals over all views. Theoretically, adding bias $F^\nu$ could produce a more accurate fitting model, which means a model has smaller residual $r$. We examine this property by using rank $s = 1, ..., c-1$. As we can see from Figure 1, for all the 4 datasets, the residual using bias is always smaller than the residual without bias using all different ranks. In Figure 1a, 1c, and 1d, the residual with bias ("With bias" line) is smaller than the residual without bias ("No bias" line). For MSRC data, the residual with bias is about 3 less than the residual without bias; for Caltech data, Figure 1b shows that the residual with bias is less than residual without bias; for Cornell data, the residual with bias is about 2 less on all rank numbers; for Cora data, the residual with bias is about 2 less on all rank numbers. In all, our results show that multi-view regression using bias could produce more accurate fitting models with less model residuals. In the following experiments, the default setting of all experiments is using bias.

**Classification using regression** In multi-view regression, there are different ways to do classification. For single-view low-rank regression (Cai et al. 2013),

$$\min_{A,B} \|Y - X^T AB\|_F^2 + \lambda\|AB\|_F^2, \qquad (19)$$

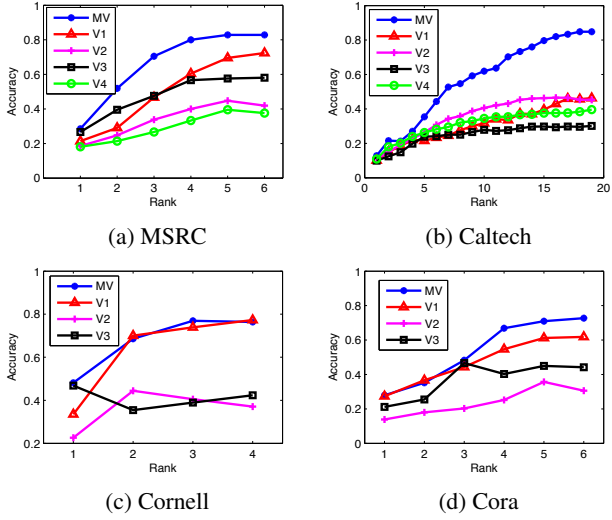

(a) MSRC      (b) Caltech

(c) Cornell      (d) Cora

Figure 4: Classification results of multi-view vs. single view data.

where $A \in \Re^{p \times s}$, $B \in \Re^{s \times c}$ and $AB$ is the low-rank regression parameter matrix, the following decision function is applied to classify a testing point $\mathbf{x} \in \Re^{p \times 1}$ into one of the $c$ classes,

$$\arg\max_{1 \leq j \leq c} (\mathbf{y})_j, \qquad (20)$$

where vector $\mathbf{y} = \mathbf{x}^T AB \in \Re^{1 \times c}$, class $j$ corresponds to the index of the maximum value in vector $\mathbf{y}$.

In multi-view case, we predict a class using each view and then use majority voting to decide the final class. For example, for view $\nu$, we use the following decision function to classify a testing point $\mathbf{x}_\nu \in \Re^{p \times 1}$ into one of the $c$ classes,

$$\arg\max_{1 \leq j \leq c} (\mathbf{y}_\nu)_j, \qquad (21)$$

where vector $\mathbf{y}_\nu = \mathbf{x}_\nu^T A_\nu B \in \Re^{1 \times c}$, $\mathbf{x}_\nu$ is the data vector of view $\nu$, $\nu = 1, 2, ..., v$. Thus we predict a class label using every view. We have $v$ predicted classes and apply majority voting on $v$ results. The class with most votes is assigned to this data point. If the top two classes get same number of votes, we assign them with 0.5 probability, etc.. We call this majority voting as "Voting" in Figure 2.

In this regression prediction problem, however, we can theoretically *derive* another voting method denoted as "Sum". Since our starting point is Eqs.(1-3), after obtaining $A_\nu$ and $B$ through training, for a testing point $\mathbf{x}$, we learn $\mathbf{y}$ that minimizes the difference between label vector $\mathbf{y}$ and projected data of each views $\mathbf{x}_\nu^T A_\nu B$:

$$\min_{\mathbf{y}} \sum_{\nu=1}^{v} \|\mathbf{y} - \mathbf{x}_\nu^T A_\nu B\|_F^2. \qquad (22)$$

It is obvious that the solution of Eq.(22) is given as

$$\mathbf{y} = (\sum_{\nu=1}^{v} \mathbf{x}^T A_\nu B)/v. \qquad (23)$$

Once $\mathbf{y}$ is computed, we use Eq.(20) to obtain the class.

The classification accuracy using the two methods, Sum and Voting, is shown in Figure 2. As we can see from the results, for data Caltech, Cornell and Cora, Sum method has better results than Voting method obviously. Overall, the Sum voting method is better for regression based classification approach for multi-view regression. In the following experiments, the default setting of every experiment is using Sum method.

**Regularization weight parameter** $\lambda_\nu$ Regularization weight parameter $\lambda_\nu$ affects the regression model and classification accuracy directly. Many researchers tune this regularization weight parameter exponentially within a specific domain, such as from $10^{-5}$ to $10^5$. It is very time consuming and misleading. In fact, regularization weight parameter $\lambda_\nu$ has direct contribution to the eigenvalues of $(X_\nu X_\nu^T + \lambda_\nu I)$, as shown in Eq.(7). A large $\lambda_\nu$ could change the distribution of eigenvalues of $(X_\nu X_\nu^T + \lambda_\nu I)$ significantly. While a small $\lambda_\nu$ preserves the original eigenvalues distribution of $X_\nu X_\nu^T$. Thus, we constrain $\lambda_\nu$ to be the following 3 cases:

1. The summation for all the eigenvalues of $X_\nu X_\nu^T$. This will change the distribution of eigenvalues of $(X_\nu X_\nu^T + \lambda_\nu I)$ more significantly. Since $X_\nu$ is normalized row-wisely, $\lambda_\nu = \text{Tr}(X_\nu X_\nu^T) = p_\nu$, where $p_\nu$ is dimension of view $\nu$. In Figure 3, result using this method is denoted as "p".

2. The average of all the eigenvalues of $X_\nu X_\nu^T$. So $\lambda_\nu = \text{Tr}(X_\nu X_\nu^T)/p_\nu = 1$, where $p_\nu$ is dimension of view $\nu$. In Figure 3, result using this method is denoted as "1".

3. The 90%th largest eigenvalue. For example, if $X_\nu X_\nu^T$ has 200 non-zero eigenvalues sorted from large to small, we let $\lambda_\nu$ be the $90\% \times 200 = 180$th eigenvalue. This will change the distribution of eigenvalues of $(X_\nu X_\nu^T + \lambda_\nu I)$ slightly and still preserve the original eigenvalue distribution of $X_\nu X_\nu^T$. In Figure 3, result using this method is denoted as "p90".

Figure 3a shows that, for MSRC data, $\lambda_\nu = 1$ and "p90" performs better than using the summation of all eigenvalues ($\lambda_\nu = p_\nu$). In Figure 3b, $\lambda_\nu = 1$ can beat "p90" and $\lambda_\nu = p_\nu$. In Figure 3c, $\lambda_\nu = 1$ also has the best accuracy for rank $s = 2, 3, 4$. For data Cora, using different $\lambda_\nu$ does not affect accuracy too much. Over all, we choose $\lambda_\nu$ as the average of all eigenvalues of $X_\nu X_\nu^T$, which is $\lambda_\nu = 1$. In the following experiments, the default setting of every experiment is using $\lambda_\nu = 1$.

**Comparison with single view**

Multi-view regression uses data or information from multiple channels, such as different image features, both webpage citations view and contents view. Generally, we expect that multi-view regression can produce better results by exploiting information from multiple views. In this part, we compare multi-view low-rank regression with single-view low-rank regression (see (Cai et al. 2013)). Figure 4 shows that multi-view low-rank regression produces better classification accuracy than single-view regression for different ranks (rank $s$ is from 1 to $c - 1$). "MV" denotes multi-view accuracy, "V1", "V2", ..., denote the accuracy using different single view. For example, Figure 4a shows that, for data MSRC, multi-view regression has much higher accuracy than all single-view low-rank regression when rank $s = 2, 3, 4, 5, 6$. Figure 4b shows that, when rank $s > 4$, multi-view regression has much higher accuracy than all the four single views. In Figure 4c, view "V1" has very good accuracy, but multi-view regression has better results than view "V1" when $s = 1, 3$. In Figure 4d, multi-view outperforms single-view when $s = 4, 5, 6$.

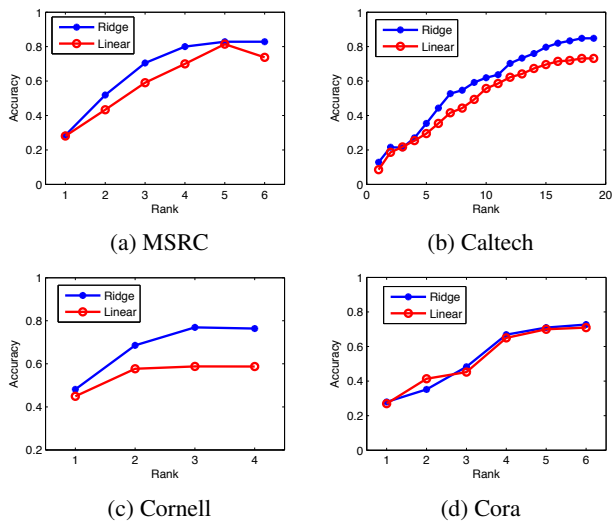

Figure 5: Comparison of ridge regression and linear regression.

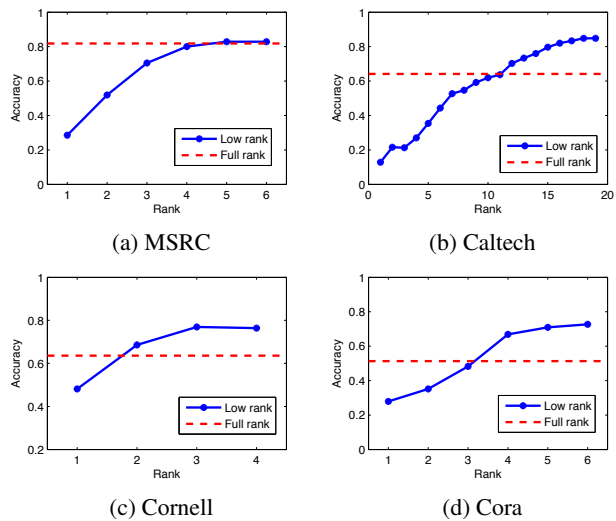

Figure 6: Comparison of low-rank and full-rank.

## Comparison of ridge regression and linear regression

Linear regression (when $\lambda_\nu = 0$) and ridge regression (when $\lambda_\nu \neq 0$) are closely related. Previous research (Hoerl and Kennard 1970) (Cai et al. 2013) shows that ridge regression will have better performance than linear regression. However, all existing work is based on single view. Does multi-view ridge regression produce better results than multi-view linear regression? We will examine the performance of multi-view linear regression and ridge regression on the 4 multi-view data with respect to different ranks. We can get linear regression by simply setting $\lambda_\nu = 0$ in our existing multi-view ridge regression model. Figure 5 shows that multi-view low-rank **ridge** regression ("Ridge" line in the figure) produces better classification accuracy than multi-view low-rank **linear** regression (when $\lambda_\nu = 0$, "Linear" line in the figure) in datasets MSRC, Caltech and Cornell. For dataset Cora, ridge regression get slightly better results than linear regression when rank $s = 3, 4, 5, 6$.

## Comparison of low-rank and full-rank

In real life, low-rank reveals the underlying structure of datasets and removes the noise and redundant information in the datasets. Low-rank regression model has been proved to be an effective learning mechanism by exploring the low-rank structure of real life data (Xiang et al. 2012) (Evgeniou and Pontil 2007) (Cai et al. 2013). For full-rank regression, there is no constraint on $B^\nu$ in Eq.(1). We minimize the objective function of full-rank regression Eq. (16) and use the closed-form optimal solution given by Eq.(17) to solve the full-rank objective.

Figure 6 compares classification accuracy using low-rank multi-view regression and full-rank multi-view regression. The blue dot line is the low-rank classification accuracy for rank $s = 1, ..., c - 1$, where $c$ is class number. The red dash line is full-rank classification accuracy with rank $s = c$. The horizontal axis denotes rank of regression and the vertical axis denotes classification accuracy. As we can see, for all the 4 datasets, low-rank regression model can always beat full-rank regression model. For example, in Figure 6a, low-rank results with $s = 5$ and $s = 6$ have higher accuracy than full-rank with $s = 7$ (red dash line). In Figure 6b, low-rank results with $s = 11$ to $s = 19$ have higher accuracy than full-rank with $s = 20$. Figure 6c shows low-rank results with $s = 2, 3, 4$ have higher accuracy than full-rank with $s = 5$. Figure 6d shows low-rank results with $s = 4, 5, 6$ have higher accuracy than full-rank with $s = 7$.

## Conclusion

In this paper, we proposed a multi-view low-rank regression model. We provide a closed-form solution to multi-view low-rank regression model. Extensive experiments conducted on 4 multi-view datasets show that multi-view low rank regression outperforms full-rank regression counterpart and single-view counterpart in terms of classification accuracy.

**Acknowledgement**. Research is partially supported by NSF-IIS 1117965, NSF-IIS 1302675, NSF-IIS 1344152, NSF-DBI 1356628.